# Multi-Fold Gabor, PCA and ICA Filter Convolution Descriptor for Face Recognition

Cheng-Yaw Low, Andrew Beng-Jin Teoh, *Senior Member*, *IEEE*, Cong-Jie Ng

*Abstract*—This paper devises a new means of filter diversification, dubbed multi-fold filter convolution ($\mathcal{M}$-FFC), for face recognition. On the assumption that $\mathcal{M}$-FFC receives single-scale Gabor filters of varying orientations as input, these filters are self-cross convolved by $\mathcal{M}$-fold to instantiate a filter offspring set. The $\mathcal{M}$-FFC flexibility also permits cross convolution amongst Gabor filters and other filter banks of profoundly dissimilar traits, e.g., principal component analysis (PCA) filters, and independent component analysis (ICA) filters. The 2-FFC of Gabor, PCA and ICA filters thus yields three offspring sets: (1) Gabor filters solely, (2) Gabor-PCA filters, and (3) Gabor-ICA filters, to render the learning-free and the learning-based 2-FFC descriptors. To facilitate a sensible Gabor filter selection for $\mathcal{M}$-FFC, the 40 multi-scale, multi-orientation Gabor filters are condensed into 8 elementary filters. Aside from that, an average histogram pooling operator is employed to leverage the 2-FFC histogram features, prior to the final whitening PCA compression. The empirical results substantiate that the 2-FFC descriptors prevail over, or on par with, other face descriptors on both identification and verification tasks.

## I. Introduction

FACE recognition, including identification and verification tasks, is highly challenging in practice due to wide intra-class variability in pose and expression, and other disturbances, including illumination, occlusion, misalignment, corruption, just to name a few. An ideal face descriptor, regardless of handcrafted or learning-based, should be invariant to these intra-class difficulties. A plausible remedy is the longstanding filter bank (FB) approaches, where the local structure of overlapping neighborhoods is featured by means of linear local convolutions, or local matches [1], [2]. This is, in general, in line with the renowned deep convolutional neural network (CNN) models [3-5] from the feature extraction perspective. However, the key success factor of the CNNs relies on the availability of large scale training datasets, e.g., DeepFace by Facebook [3] demands a gigantic dataset of 4.4M images with over 4,000 identities; DeepID3 [4] by the Chinese University of Hong Kong learns from approximately 300,000 images with 13,000 identities; FaceNet [5] by Google trains CNNs from 200M images spanning over 8M identities. These prevailing CNN models, particularly DeepID3 and FaceNet, reportedly achieve accuracies of 99.53% and 99.63%, respectively, on the labeled faces in the wild (LFW) dataset [41], surpassing the human-level performance of 97.53%. On the contrary, the FB approaches, e.g., PCANet [14], discriminant face descriptor (DFD) [15], compact binary face descriptor (CBFD) [16], binarized statistical image features (BSIF) [17-18], DCTNet [20], etc., are typically equipped with a single or two filtering layers. Despite of being simple and easy of use, these CNN simplifications promise the state of the art robustness to the generic image classification problems including face.

The earliest FB approaches are reviewed and compared in [6]. They share a common three-stage pipeline, referred to as filter-rectify-filter (FRF): (1) a **convolutional** stage based on the heuristically designed filter banks, e.g., Laws masks, ring and wedge filters, Gabor filters, wavelet transform, packets and frames, discrete cosine transform (DCT), etc.; or other optimal filters, e.g., principal component analysis (PCA) eigenfilters, Karhunen-Loeve transform, prediction error filters, optimized Gabor filters, etc., (2) a **nonlinearity**, a. k. a filter response rectification step, e.g., magnitude, squaring, rectified sigmoid, etc., (3) **pooling** (filtering) operations, e.g., spatial averaging, smoothing, or nonlinear inhibition, to remove the inhomogeneity in the rectified responses within a homogenous region. The local energy function, includes stage (2) and (3), outputs a set of feature images, one per filter, defining the bases for classification.

In lieu of local energy estimation, Unser [2] summarizes the $N$th-order probability density function (PDF) of the filter responses by $N$ histograms. Liu et al. [7-8] streamlines the Unser's work by only retaining the first-order PDF, denoted by *spectral histogram* (SH). This is supported by the statistical ground that, on the independence assumption, the underlying joint probability distribution of the local neighborhoods is characterized by the marginal PDFs, and that the SH is of lower-complexity PDF approximation. The $\chi^2$–statistic, on the other hand, is laid out in [7] as a distance metric comparing two SHs, and the important SH properties are provided in [8].

The data-driven paradigm is first incorporated into the SH techniques in [9-10], where a vocabulary of prototype, a. k. a. codebook, is constructed using the unsupervised $k$-means algorithm. The filter responses are afterward associated to the nearest codeword for the convenience of histogram formation. The emergence of the local binary pattern (LBP) [11] texture

C. Y. Low, A. B. J. Teoh, and C. J. Ng are now with the School of Electrical and Electronic Engineering, College of Engineering, Yonsei University, Seoul, South Korea E-mail: {chengyawlow, bjteoh, congjie}@yonsei.ac.kr.

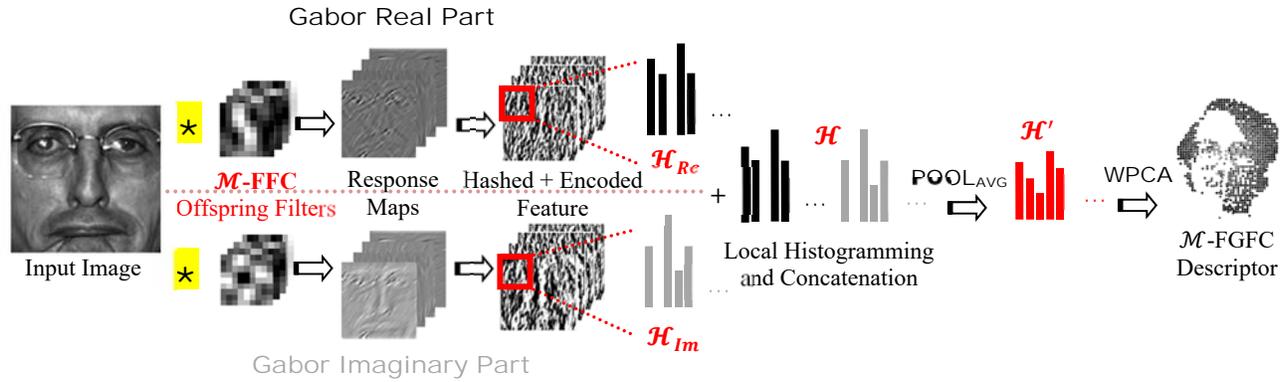

Fig. 1. The generic 3-stage $\mathcal{M}$-FFC framework in Gabor domain with real and imaginary constituents: (1) image-filter convolution, (2) binarization and feature encoding, (3) local histogramming and concatenation, followed by an average pooling unit and WPCA. The offspring filters are either the $\mathcal{M}$-FFC outcome of Gabor filters, Gabor and PCA filters, or Gabor and ICA filters.

operator leads to another distinguished SH variant. Ahonen and Pietikäinen [12] equalize the LBP derivation to the FB implementation whereby the LBP is interpreted as a filtering operator based on a set of local derivative filters. The tight connection between the LBP and the FB approaches allows the filter responses to be zero-thresholded and the LBP feature encoding is simulated to denote the extracted features in the SH representation. Heikkilä and Pietikäinen [13] observe from experiments that, under some circumstances, a non-zero threshold might be chosen to stabilize the binary codes.

### A. Related Works

In comparison to the FRF model, the contemporary 3-stage SH pipeline, as in [12], is of more parsimonious: (1) a **convolutional** stage, where the present face recognition works apply Gabor, DCT, or the pre-learned filters, particularly PCA filters, independent component analysis (ICA) filters, and linear discriminant analysis (LDA) filters, (2) a **non-linearity**, i.e., hashing (binarization) and LBP-alike feature encoding, (3) a **pooling** operation, i.e., histogramming. One of the recent SH techniques is PCANet [14], where the PCA filters are learned from the 2-layer net topology. The experimental results unveil that the PCANet performance remarkably prevails over other state of the arts, including the LDA-learned face descriptors, i.e., DFD [15], and CBFD [16].

In contrast to PCANet, BSIF [17-18] train the single-layer ICA filter banks from 13 natural images and face regions, such that the statistical independency amongst the ICA responses is maximized. The BSIF performance reported in [18] attests that the block-wise ICA filters stemmed from varying face regions, in conjunction with the Tan and Triggs's (TT) photometric normalization [19], outperform those learned from the pre-selected natural images on the FERET and the LFW datasets. Inspired by PCANet and BSIF, DCTNet is outlined by Ng and Teoh [20]. Despite of being learning-free, the DCTNet performance is shown to be on par with PCANet in the face identification task, particularly in the pose-oriented cases.

Other relevant face descriptors are Gabor-based derivatives, e.g., [23-28] (to be detailed in Section II), LBP variants, e.g., [29-31], and CNN-motivated models, e.g., [32-34]. With these exertions, the traditional face recognition problem is re-defined, shifting from the strictly regulated setting to the unconstrained condition with severe intra-variabilities, e.g., the LFW and the YouTube Faces (YTF) images. This evolvement stimulates enormous research on pose-invariant face recognition, e.g., [35-37].

### B. Motivation

This work is mainly motivated by PCANet, BSIF, DCTNet and other relevant variations. We discern from our experiments that the multi-layer PCANet is incompetent to abstract high-level filters like the CNN models do. In other words, the learning capability of PCANet is only limited to that of single-layer, where it only learns very similar low-level filters in every layer (to be revealed in Section III (F)). The PCANet architecture is thus inappropriate to be seen as a deep learning network instance.

Another major shortcoming of PCANet and BSIF is that, the amount of learnable PCA and ICA filters depends on the local patch dimension directly. For the local patches of $5 \times 5$, as an example, there are at most 25 learnable PCA and ICA filters. One can expand the local patch dimension to learn more PCA and ICA filters. However, only the top-ranked PCA and ICA filters are utilized for feature extraction. For example, PCANet only uses 8 PCA filters in the first and the second layers, while BSIF fixes the number of ICA filters to 11 in the experiments. We, therefore, term this situation as *filter scarcity*, where the limited filters are deficient to extract discriminative features, especially from the unconstrained images. Moreover, the CNN models, from DeepFace to the recent FaceNet, recruit a huge number of filters, in addition to being deep.

To cope with the filter scarcity issue, we propose multi-fold filter-to-filter convolution, referred to as $\mathcal{M}$-FFC hereinafter, to exponentially proliferate the convolution filters, including PCA and also ICA filters, in the complex Gabor domain. We conduct extensive experiments to prove the direct benefit of $\mathcal{M}$-FFC in feature extraction in Section IV (D).

### C. Contribution

The main contribution of this paper is three-fold: (1) a new means of *filter diversification* via $\mathcal{M}$-FFC is devised, where the Gabor filters are either *self-cross convolved* by $\mathcal{M}$-fold, or with other pre-learned PCA and ICA filters to instantiate the offspring sets; (2) the standard Gabor filter bank of 5 scales and 8 orientations in face recognition are recapitulated into the condensed Gabor filter ensemble of only 8 elementary filters, abbreviated as **Gabor**$_{Std}$ and **Gabor**$_{Cond}$, respectively, in this paper; (3) the $\mathcal{M}$-FFC histogram features are leveraged by an average pooling unit to mitigate the dimensionality problem. The 3-stage $\mathcal{M}$-FFC pipeline, as portrayed in Fig. 1, is similar to other SH techniques but in the complex Gabor domain. This signifies that each $\mathcal{M}$-FFC operation yields two offspring sets: one for each real and imaginary part, for feature extraction.

This paper, in general, builds on our previous work in [56], where we offers a more comprehensive analysis and a detailed exposition in terms of the $\mathcal{M}$-FFC flexibility and variations. In addition to the learning-free $\mathcal{M}$-FFC$_{Gabor}$ descriptor reported earlier (based on Gabor filters alone), the learning-based $\mathcal{M}$-FFC descriptors, i.e., $\mathcal{M}$-FFC$_{Gabor-PCA}$, and $\mathcal{M}$-FFC$_{Gabor-ICA}$ (based on Gabor, PCA and ICA filters), are also explored. We conduct extensive experiments to study the performance of the $\mathcal{M}$-FFC descriptors, not limited to face identification, but also the face verification task on the challenging LFW and the YTF datasets.

The remainder of this paper is organized as follows: Section II underlines the preliminary parts of this work, and Section III elaborates the $\mathcal{M}$-FFC pipeline. What follows is the $\mathcal{M}$-FFC performance analysis and discussion in Section IV, and the concluding remark is provided in the last section.

## II. PRELIMINARY

In this section, Gabor Wavelets and some Gabor-related works in face recognition are reviewed, followed by the PCA and ICA filter learning practiced in our experiments.

### A. Gabor Wavelets

Gabor wavelets [21], a filter repository pre-tuned to various spatial frequencies (scales) and orientations, have sparked off much novel research ideas in face recognition. In the spatial domain, the 2D Gabor wavelets in $v$ scales and $u$ orientations are expressed as follows:

$$\psi_{u,v}(z) = \frac{\|k_{u,v}\|^2}{\sigma^2} e^{(-\|k_{u,v}\|^2 \|z\|^2 / 2\sigma^2)} \left[ e^{ik_{u,v}z} - e^{-\sigma^2/2} \right] \quad (1)$$

where $z = (x, y)$, $\sigma$ denotes the Gabor envelop width, and $k_{u,v} = k_v e^{i\phi_u}$ refers to the wavelet vector with $k_v = k_{max}/f^v$ and $\phi_u = u\pi/8$. The Gabor-related face recognition works found in the literature typically configure the free parameters as: $u \in \{0, \ldots, 7\}$, $v \in \{0, \ldots, 4\}$, $\sigma = 2\pi$, $k_{max} = \pi/2$, and $f = \sqrt{2}$ [23], [25], [27-28]. Therefore, this parameter setting is termed the *de factor* standard in this paper, where it derives **Gabor**$_{Std}$ of 40 multi-scale, multi-orientation filters for each real and imaginary part. The Gabor response of an arbitrary image $I$ is described as the convolution of $I$ and every Gabor filter $\psi_{u,v}$ as follows:

$$\mathcal{G}_{u,v}(z) = I(z) * \psi_{u,v}(z) \quad (2)$$

where $z = (x, y)$, and * indicates the convolution operator.

The Gabor-Fisher classifier (GFC) introduced by Liu and Wechsler [23] is one of the earliest Gabor face representation works. In place of SH, the enhanced Fisher linear discriminant (FLD) is applied to the augmented Gabor responses to generate the low-dimensional features. The most well-known Gabor-based SH exemplars are Zhang et al. [24] and Zhang et al. [25]: the former transforms the Gabor magnitude pictures to the local Gabor binary patterns by using the LBP operator, and the latter adopts the local XOR pattern (LXP) operator to manipulate the binarized Gabor responses, before forming the respective histogram features. Xie et al. [26] is an extension to [25], where the block-based FLD is used to compress the Gabor phase and magnitude features. On the contrary, Lei et al. [27] concatenates the histogram features in spatial, scale and orientation domains, followed by the LDA-based discriminant classification. Similar to [9] and [10], Hussain [28] learns a codebook to histogram the Gabor filter responses. Most of the existing Gabor-based SH variants necessitates a non-linearity, either LBP, or LXP, to achieve satisfactory performance. To the best of our knowledge, no work is conducted on defining the histogram features directly from the Gabor responses. An exception is the GGPP component presented in [25], where it is aggregated with the histogram features generated by the local XOR patterns as the final descriptors.

### B. PCA and ICA Filter Ensemble Learning

Suppose that we are to build the data-driven PCA and ICA filter ensembles from $M$ training images $\{I_m \in \mathcal{R}^{h \times w}\}_{m=1}^{M}$. The local pixel-wise patches of size $k \times k$ are identified, and the zero-mean patches are obtained by removing the mean intensity from each patch. A subset of $N$ random patches are selected and stacked in columns as follows:

$$I' = [I'_1, I'_2, \ldots, I'_N] \in \mathcal{R}^{k^2 \times N} \quad (3)$$

Assuming that we target to learn $i$ PCA filters from $I'$ (to be recruited in turn to define $i$ ICA filters), the $i$ orthonormal PCA bases $W \in \mathcal{R}^{k^2 \times i}$ are estimated by minimizing the reconstruction error as follows:

$$\min_{W \in \mathcal{R}^{k^2 \times i}} \| I' - WW^T I' \|^2, s.t. W^T W = \mathbb{I}_i \quad (4)$$

We thus furnish the PCA filter ensemble $W_{PCA}$ with $i$ principal eigenvectors extracted from the eigensolution for $I'(I')^T$ as:

$$W_{PCA} = [w_{PCA,1}, w_{PCA,2}, \ldots, w_{PCA,i}]^T \in \mathcal{R}^{i \times k^2} \quad (5)$$

where $W_{PCA}$ captures the greatest $i$ variations of $I'$.

Let $W_{ICA} \in \mathcal{R}^{i \times k^2}$ be the ICA filter ensemble to be learned from $I'$ by maximizing the statistical independent of $\mathcal{Z} \in \mathcal{R}^{i \times N}$; $W_{ICA}$ is decomposed into two parts:

$$W_{ICA} = UV \quad (6)$$

$$\mathcal{Z} = W_{ICA} I' = UV I' \quad (7)$$

where $U \in \mathcal{R}^{i \times i}$ is an ICA estimation, and $V \in \mathcal{R}^{i \times k^2}$ is the whitening transformation matrix performing the canonical processing based on $W_{PCA}$. Mathematically, $V$ is defined as:

$$V = D^{-1/2} \cdot W_{PCA} \quad (8)$$

where $D \in \mathcal{R}^{i \times i}$ describes the sorted $W_{PCA}$ eigenvalues in the main diagonal entries. Assuming also that $VI' \in \mathcal{R}^{i \times N}$ is the dimension-reduced and whitened patches, the orthogonal $U$ is estimated based on the fast ICA algorithm [38] to yield the independent components $Z$. Pursuant to (6), the ICA filter ensemble is filled with $W_{ICA}$ based on $U$ and $V$.

To sum up, the PCA and ICA filter ensemble learning takes 3 primary stages: (1) a set of $N$ mean-removed local patches are sampled from training images; (2) the PCA filter ensemble is obtained from the foremost $i$ principal eigenvectors, and the PCA dimension reduction and whitening are exercised on the local patches; (3) The ICA filters are estimated with respect to the dimension-reduced and whitened patches.

### III. $\mathcal{M}$-FFC Face Descriptor Formulation

In this section, the $\mathcal{M}$-FFC pipeline is detailed and compared to the most relevant counterparts. The computational complexity of each is also analyzed.

#### A. Condensed Gabor Filter Ensemble

The typical Gabor-based SH techniques [24-28] suffer from the curse of dimensionality issue as a consequence of stacking all histogram features for Gabor magnitude or/and phase. In lieu of that, this paper condenses the 40 standard multi-scale, multi-orientation Gabor filters in face recognition, **Gabor**<sub>Std</sub>, by averaging the Gabor filters of the same orientation across varying scales to define **Gabor**<sub>Cond</sub> as follows:

$$\Phi = \left\{ \varphi_f = \frac{1}{v_{max}} \sum_{v=0}^{v_{max}-1} \psi_{u,v}, f = u + 1 \right\}_{u=0}^{u_{max}-1}, \quad (9)$$
$$\psi_{u,v}, \varphi_f \in R^{k \times k}$$

where $u_{max}$ and $v_{max}$ are set to 8 and 5, respectively, in this paper. Fig. 3(a) pictorializes that the resultant 8 elementary **Gabor**<sub>Cond</sub> filters are in the initial 8 orientations. On top of that, the frequency spectrum portrayed in Fig. 2 shows that, despite of only 8 filters, the **Gabor**<sub>Cond</sub> set envelops the entire **Gabor**<sub>Std</sub> frequency space. We demonstrate in Section IV (C) that the **Gabor**<sub>Cond</sub> performance is comparable to that of **Gabor**<sub>Std</sub>, with negligible degradation.

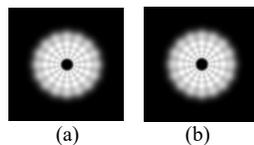

Fig. 2. Frequency spectrum for **Gabor**<sub>Std</sub> of 40 filters (a), and **Gabor**<sub>Cond</sub> of only 8 filters (b).

#### B. $\mathcal{M}$-FFC Filter Diversification

For the most basic $\mathcal{M}$-FFC instance, let the **Gabor**<sub>Cond</sub> filter set in (9) be the $\mathcal{M}$-FFC input, $\Phi^{(m)} = \left\{ \varphi_f^{(m)} \in R^{k \times k} \right\}_{f=1}^{\mathcal{F}_m}$, given that $m = 1, \ldots, \mathcal{M}$; these filters are *self-cross convolved* with one another to render an Gabor offspring set $\mathcal{O}_{\text{Gabor}}$ of $\mathcal{L} = \prod_{m=1}^{\mathcal{M}} \mathcal{F}_m$ filters as follows:

$$\mathcal{O} = \left\{ \sigma_\ell = \varphi_{1,\ldots,\mathcal{F}_1}^{(1)} * \varphi_{1,\ldots,\mathcal{F}_2}^{(2)} * \ldots * \varphi_{1,\ldots,\mathcal{F}_\mathcal{M}}^{(\mathcal{M})} \right\}_{\ell=1}^{\mathcal{L}} \quad (10)$$

where $\mathcal{O} = \{\mathcal{O}_{\text{Gabor}}, \mathcal{O}_{\text{Gabor-PCA}}, \mathcal{O}_{\text{Gabor-ICA}}\}$, $\sigma_\ell \in R^{\mathcal{K} \times \mathcal{K}}$ and $\mathcal{K} = \mathcal{M}(k-1) + 1$. This implies that, for $\mathcal{M} = 2$, $\mathcal{F}_1 = \mathcal{F}_2 = 8$, and $k = 3$, where the two folds are fed with the same eight **Gabor**<sub>Cond</sub> filters of each $3 \times 3$, the $\mathcal{O}_{\text{Gabor}}$ of $\mathcal{L} = 64 (= 8 \times 8)$ offspring of each $5 \times 5 (= 2(3-1)+1)$ is produced as a consequence of the 2-FFC filter diversification; for $\mathcal{M} = 3, \mathcal{F}_1 = \mathcal{F}_2 = \mathcal{F}_3 = 8$, and $k = 3$, the 3-FFC operation outputs $\mathcal{L} = 512 (= 8 \times 8 \times 8)$ offspring of each $7 \times 7 (= 3(3-1)+1)$. To retain the original filter size of $k \times k$, one can simply return the central part of the $\mathcal{M}$-FFC output.

Fig. 3 illustrates the 2-FFC filter diversification progression based on the 8 **Gabor**<sub>Cond</sub> filters. For simplicity, the resultant $\mathcal{O}_{\text{Gabor}}$ of $\mathcal{L} = 64$ offspring is pruned to $36 (= (8+1) \times 8/2)$ unique filters, as the convolution operation is of *commutative* (applicable to $\mathcal{O}_{\text{Gabor}}$ only, due to filter redundancies). We also notice from Fig. 3 that the offspring filters are implanted with distinctive characteristics, e.g., the checkerboard-alike filters are obtained by convolving two filters of different directions. The offspring set is therefore capable of featuring the salient textures consistent to the filter structures, on top of being sensitive to the local edges (just like the ordinary 1-fold Gabor filters). In other words, the offspring serve dual role: edge and texture detectors. This is empirically examined in Section IV.

The $\mathcal{M}$-FFC paradigm is inherently flexible. Apart from the Gabor-Gabor setup, the $\mathcal{M}$-FFC flexibility also permits filter convolution amongst **Gabor**<sub>Cond</sub> and other filter banks, e.g., the learning-based PCA and ICA filters, to complement the learning-free **Gabor**<sub>Cond</sub> for the discriminant feature extraction stage. To remain in the complex Gabor domain, for the 2-FFC instance, the pre-learned PCA and ICA filters are received as the fold-2 input. This derives the Gabor-PCA and Gabor-ICA offspring sets, hereinafter referred to as $\mathcal{O}_{\text{Gabor-PCA}}$ and $\mathcal{O}_{\text{Gabor-ICA}}$, respectively. The $\mathcal{O}_{\text{Gabor-PCA}}$ and $\mathcal{O}_{\text{Gabor-ICA}}$ offspring sets are pictorialized in Fig. 4 and 5. The PCA and ICA filter learning steps and the implementation details are respectively given in Section II (B). Note that, the ideal PCA orthonormal and the ICA independent properties are destructed due to 2-FFC. We demonstrate in Section IV (C) that this expense is repaid with a substantial improvement in feature discriminability.

In summary, the $\mathcal{M}$-FFC operation produces $\mathcal{L}$ offspring for each Gabor real and imaginary constituent. In practice, $\mathcal{M}$ is empirically determined for certain degree of performance gain resulted from filter diversification. The $\mathcal{M}$-FFC demerit is that $\mathcal{L}$ rises exponentially with respect to $\mathcal{M}$. This paper, however, only restricts the exploration to 1-FFC and 2-FFC.

#### C. Convolutional Stage

Provided with an image $I \in R^{h \times w}$, and an $\mathcal{M}$-FFC offspring set $\mathcal{O} = \{ \sigma_\ell \in R^{\mathcal{K} \times \mathcal{K}} \}_{\ell=1}^{\mathcal{L}}$, $I$ is convolved with $\sigma_\ell$ to yield $\mathcal{L}$ responses as follows:

$$G = \{ g_\ell = I * \sigma_\ell \}_{\ell=1}^{\mathcal{L}}, g_\ell \in R^{h \times w} \quad (11)$$

where $I$ is zero-padded by $(\mathcal{K}-1)/2$ in advance before the convolution operation to remain the filter response size similar to $I$. If $I$ is represented by columnar zero-mean local patches $I' \in R^{\mathcal{K}^2 \times hw}$, (11) is expressed as a linear matrix projection of $\mathcal{O}'$ and $I'$:

$$G' = \mathcal{O}' I' \quad (12)$$

where $\mathcal{O}' \in R^{\mathcal{L} \times \mathcal{K}^2}$ is a matrix stacked with the $\mathcal{L}$ offspring in rows, and $G' \in R^{\mathcal{L} \times hw}$ is a composite response matrix to be re-organized into $G$ accordingly.

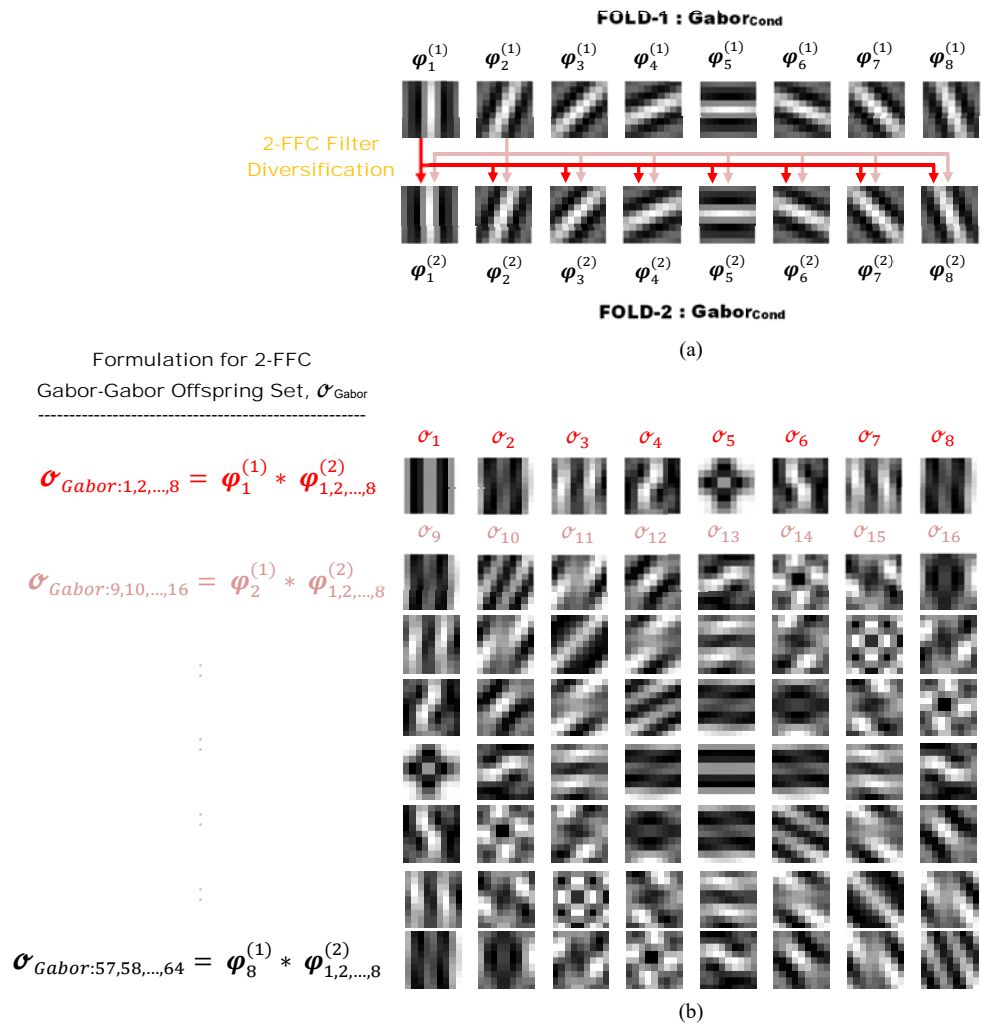

Fig. 3. Let fold-1 and fold-2 accommodate **Gabor$_{Cond}$** of 8 elementary filters each; these filters are self-cross convolved via 2-FFC to derive 2-FFC Gabor-Gabor offspring set $\mathcal{O}_{Gabor}$ of 36 unique filters for subsequent discriminative feature extraction stage.

### D. Binarization, Feature Encoding, and Histogramming

The non-linear hashing operator thresholds the $\mathcal{L}$ filter responses, refer to $\mathcal{G}$ in (11), with respect to 0 to assign a bit '1' to the positive coefficients, and a bit '0' otherwise. In what follows, the binarized filter responses are decimalized into the $\mathcal{F}_\mathcal{M}$-bit integers, ranging from 0 to $2^{\mathcal{F}_\mathcal{M}} - 1$, to define $\mathcal{D}^{(t)}$ as follows:

$$\mathcal{D}^{(t)} = \sum_{f=1}^{\mathcal{F}_\mathcal{M}} \mathcal{S}\big(\mathcal{G}_{(t-1)\times\mathcal{F}_\mathcal{M}+f}\big) 2^{f-1}, \quad t = 1, \dots, T \quad (13)$$

where $T = \prod_{m=1}^{\mathcal{M}-1} \mathcal{F}_m$ and $\mathcal{S}(\cdot)$ corresponds to the Heaviside step function. Subsequent to that, $\mathcal{D}^{(t)}$ is regionalized into $\mathcal{B}$ non-overlapping blocks, unless stated otherwise, where $b = 1, \dots, \mathcal{B}$. For each $b$ block, the statistical co-occurrences associated to the $2^{\mathcal{F}_\mathcal{M}}$ bins, i.e., the block-wise histograms, are aggregated as follows:

$$\hbar_b^{(t)} = \sum_{x,y} \delta\big(\alpha, \mathcal{D}_b^{(t)}(x,y)\big), \alpha = 0, \dots, 2^{\mathcal{F}_\mathcal{M}-1}, \quad (14)$$
$$\hbar_b^{(t)} \in R^{2^{\mathcal{F}_\mathcal{M}}}$$

where $\delta(\cdot)$ refers to the Kronecker delta function. The global histogram features to be employed as the $\mathcal{M}$-FGFC descriptor is a concatenation of $T.\mathcal{B}$ block-wise histograms as follows:

$$\mathcal{H} = \big[\hbar^{(t=1)}, \dots, \hbar^{(t=T)}\big] \in R^{2^{\mathcal{F}_\mathcal{M}} \mathcal{B} T} \quad (15)$$

where $\hbar^{(t)} \in R^{2^{\mathcal{F}_\mathcal{M}} \mathcal{B}}$. Since the $\mathcal{M}$-FFC framework is in the complex Gabor domain, the global histogram features for each real and imaginary part, referred to as $\mathcal{H}_{Re}$ and $\mathcal{H}_{Im}$ in Fig. 1, are cascaded as in (16) to define the holistic representation $\mathcal{H}$.

$$\mathcal{H} = [\mathcal{H}_{Re}, \mathcal{H}_{Im}] \in R^{2^{\mathcal{F}_\mathcal{M}} \mathcal{B} T 2} \quad (16)$$

In accordance to the Daugman's phase-quadrant demodulation code in [25], for the histogram features formulated based on $\mathcal{O}_{Gabor}$, the composition of $\mathcal{H}_{Re}$ and $\mathcal{H}_{Im}$ describes the Gabor phase representation. As the $\mathcal{H}$ definition in (16) unpleasantly doubles the feature dimension, an average histogram pooling operator is introduced in the succeeding section to counter the dimensionality issue.

### E. Average Histogram Pooling

**POOL$_{Avg}$**, as the name suggests, is an average pooling unit. It serves an important role of down-sampling and regulating the histogram features $\mathcal{H}$ in (16). In another word, it acts as a soft regularizer, in addition to dimension reduction, to uniformize the histogram disparity including burstiness and sparseness. It works as follows to compress $d$-dimensional $\mathcal{H}$ into $\mathcal{H}'$ of $d'$ dimensions:

$$\mathcal{H}'_d = \frac{1}{P} \sum_{n=1}^{P} \mathcal{H}_{(d-1) \times S + n}, d = 1, \ldots, D' \quad (17)$$

where $d = 2^{\mathcal{F}_\mathcal{M}} \mathcal{B} T 2$, $d' = \frac{d-P}{S} + 1$, and $P$ and $S$ describe the pooling window size and the stride step, respectively. In our experiments, $P$ and $S$ are set to 2, by default, to elicit $\mathcal{H}'$ of $2^{\mathcal{F}_\mathcal{M}} \mathcal{B} T$ dimensions. Subsequent to that, $\mathcal{H}'$ is square-rooted, L2-normalized and WPCA-ed to obtain the compact $\mathcal{M}$-FGFC descriptor.

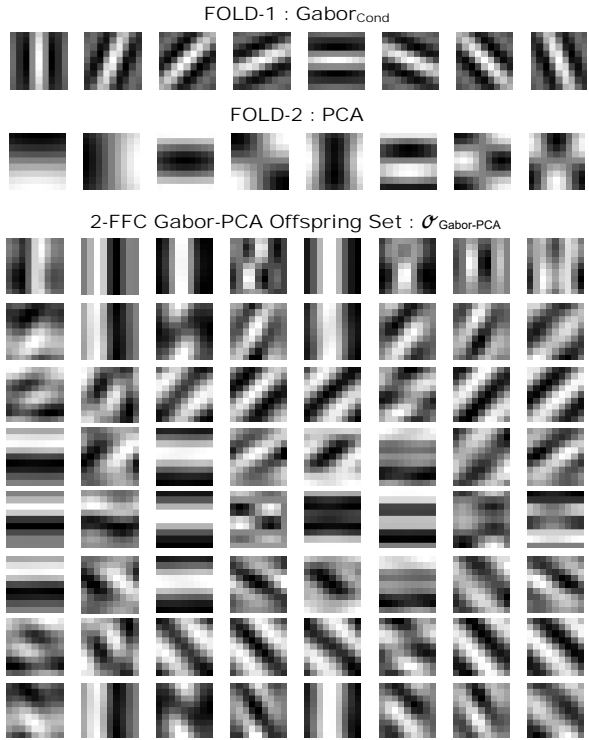

Fig. 4. 2-FFC Gabor-PCA offspring set $\mathcal{O}_{\text{Gabor-PCA}}$ derived from **Gabor**$_{\text{Cond}}$ in fold-1 and PCA filter bank in fold-2.

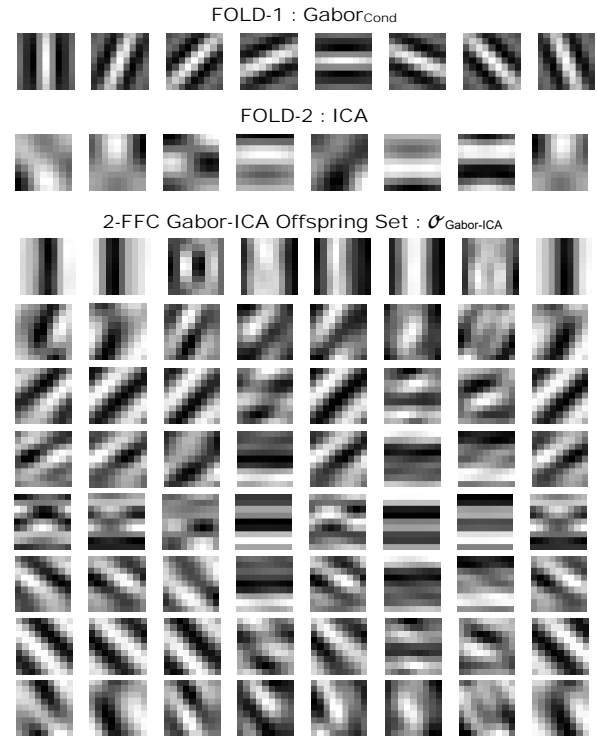

Fig. 5. 2-FFC Gabor-ICA offspring set $\mathcal{O}_{\text{Gabor-PCA}}$ derived from **Gabor**$_{\text{Cond}}$ in fold-1 and ICA filter bank in fold-2.

### F. Comparison with Relevant Techniques

The deep CNN architectures, in essence, consist of multiple interleaved convolutions, non-linear activations (e.g., absolute or square rectification, rectified linear unit, just to name a few) and max pooling (**POOL**$_{\text{Max}}$), before the last fully connected layer. As the $\mathcal{M}$-FFC backbone and other relevant techniques, e.g., PCANet [14], BSIF [17-18], and DCTNet [20], follow the 3-stage SH pipeline, they share some common grounds found in the most of CNN models. Besides the convolutional feature extraction stage, the zero-mean local patch formulation (prior to convolution stage) corresponds to the CNN local contrast normalization. Meanwhile, the most apparent discrepancy over the CNN models is that these single or 2-layer network architectures, no matter $\mathcal{M}$-FFC, PCANet, BSIF or DCTNet, contains no non-linearity, until the hashing and feature encoding stage. Since $\mathcal{M}$-FFC, BSIF, and DCTNet are of single-layer instance, the non-linear rectification layer is therefore trivial.

The PCA and ICA filter learning algorithms limit the PCA and ICA filters in the low-level form. To manifest this remark, we exercise PCANet in 5 layers using the FERE-FA images. We discern from the abstracted PCA filters in Fig. 6 that the 5-layer PCANet only learn negligibly dissimilar low-level filters in each layer. This inspires us to merge the omissible multi-layer PCANet into a flat layer, as there is no non-linearity in between the convolution layers. To be more specific, in place of learning 16 filters like the 2-layer PCANet (8 in each layer), we estimate only 8 first-layer PCA filters to be diversified based on **Gabor**$_{\text{Cond}}$ by means of 2-FFC to obtain 64 offspring. The experimental results in Section IV (D) discloses that the 2-FFC performance in face identification and verification is relatively superior to that of PCANet, BSIF, and DCTNet. This reveals that the 16 PCA filters in PCANet, the 8 layer-1 ICA filters in BSIF, and the 8 pre-fixed DCT bases in DCTNet, are insufficient to extract discriminative features in the real-world scenario. Furthermore, the CNN models involves a large set of filters in each convolution layer, e.g., FaceNet [5] learns 64 filters in the first and second layers, and the amount of filters is increased to 192, 384, and 256 in the latter layers. Therefore, the $\mathcal{M}$-FFC filter diversification is contrived to finesse the filter scarcity state, such that meaningful offspring are instantiated from multi-fold filter-to-filter convolution.

We also include a dual **POOL**$_{\text{Avg}}$ unit, where it is parallel to

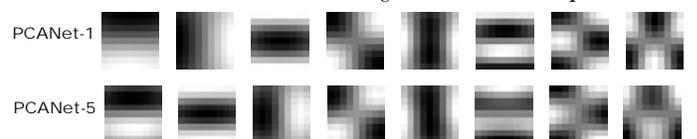

Fig. 6. PCA filter ensembles learned from 5-layer PCANet.

**POOL**$_{\text{Max}}$ in DeepFace [3], DeepID3 [4], FaceNet [5], and other representative deep CNN architectures. These operators are distinguished from one another in Section IV (C), and the empirical comparisons unveil that **POOL**$_{\text{Max}}$ is inappropriate in our case.

### G. Computational Complexity Analysis

Owing to the intense CPU consumption, the convolution operation is practically performed in the frequency domain via the fast Fourier transform (FFT). If **Gabor**$_{\text{Std}}$ is utilized for the Gabor features of an image $I_m \in \mathcal{R}^{h \times w}$, it demands $\alpha[6(hw)^2 \log_2(hw)^2 + 2(hw)^2]$ real additions, and $\alpha[4(hw)^2 \log_2(hw)^2 + 4(hw)^2]$ real multiplications, where $\alpha = u_{max} v_{max} = 8 \times 5 = 40$. On the one hand, for $\mathcal{O}_{\text{Gabor}}$ with 36 unique offspring, the cost incurred is diminished by $\alpha = 36$; for $\mathcal{O}_{\text{Gabor-PCA}}$ and $\mathcal{O}_{\text{Gabor-ICA}}$, $\alpha$ is of 64.

As regard to the single layer PCA filter learning, its time complexity is moderate. The zero-mean patches for $I'$ in (3) are derived in $k^2 + k^2 N$ flops; the covariance matrix of $I'(I')^T$ requires $2(k^2)^2 N$ flops; and the eigen-decomposition is of complexity $O((k^2)^3)$, irrelative to $N$. Overall, we expense one-half of the PCANet cost to learn $\mathcal{F}_1$ filters, equivalent to $\mathcal{F}_1 k^2$ variables.

The time complexity of fast ICA [38] running on $\mathcal{T}$ iterations is reportedly to be of $O(2\mathcal{T}k^2(k^2+1)N)$ in [54]. Although the iterative ICA estimation has no close-form solution, the cubic convergence (or at least quadratic) has been proven through the use of non-linear kurtosis contrast function and fixed-point algorithm. The fast ICA algorithm is analyzed thoroughly in terms of accuracy and computational complexity in [55].

The 2-FFC complexity for deriving $\mathcal{O}_{\text{Gabor}}$, $\mathcal{O}_{\text{Gabor-PCA}}$, and $\mathcal{O}_{\text{Gabor-PCA}}$ is of $O(\mathcal{F}_1 \mathcal{F}_2 k^2 \mathcal{K}^2)$, $k \neq \mathcal{K}$ (refer to Section III (B)). This is insignificant as $\mathcal{F}_1$, $\mathcal{F}_2$, $k$, and $\mathcal{K}$ are of small values. If $\mathcal{O}_{\text{Gabor}}$, $\mathcal{O}_{\text{Gabor-PCA}}$, and $\mathcal{O}_{\text{Gabor-PCA}}$ are computed beforehand, the computational complexity to formulate the 2-FFC descriptor for an image is as follows: the initial patch-mean removal consists of $\mathcal{K}^2 + \mathcal{K}^2 hw$ flops; the single flat convolutional layer costs $\mathcal{L}\mathcal{K}^2 hw$ flops; the hashing and feature encoding stage involves $2\mathcal{F}_2 hw$ flops; and the naive block-wise histogramming progression is of complexity $O(hw\mathcal{B}\mathcal{F}_2 \log 2)$. On the whole, the most costly part falls on the convolutional layer as $hw \gg \max(\mathcal{L}, \mathcal{K}, \mathcal{F}_2, \mathcal{B})$. In case of performing filtering in the frequency domain, the complexity is shrank to $O(\mathcal{L}hw \log(hw))$. The composite complexity is to be doubled as the 2-FFC descriptor exists in the complex domain.

## IV. EXPERIMENTS

This section underlines three 2-FFC enumerations: the learning-free 2-FFC$_{\text{Gabor}}$, and the learning-based 2-FFC$_{\text{Gabor-PCA}}$ and 2-FFC$_{\text{Gabor-ICA}}$, collectively coined 2-FFC descriptors in this work. These 2-FFC variants are rendered with respect to $\mathcal{O}_{\text{Gabor}}$, $\mathcal{O}_{\text{Gabor-PCA}}$, and $\mathcal{O}_{\text{Gabor-ICA}}$, denoting the 2-FFC offspring for 8 Gabor$_{\text{Cond}}$ filters, 8 PCA filters, and 8 ICA filters (see Fig. 3, 4, and 5). There exists another two 2-FFC possibilities, i.e., 2-FFC$_{\text{PCA}}$ and 2-FFC$_{\text{ICA}}$, where the offspring sets are the 2-FFC consequences for the pre-learned PCA and ICA filters, respectively. However, these descriptors are beyond the coverage of this paper.

We benchmark the 2-FFC performance on FERET I (frontal), FERET II (non-frontal), AR, LFW, and YouTube Faces (YTF) datasets. The experimental results are summarized in terms of rank-1 identification rate (%) for FERET I, FERET II, and AR; and area under ROC (AUC) and/or verification accuracy (%) ± standard deviation (ACC± SD) for LFW and YTF.

### A. Benchmarking Datasets

The five face datasets employed in our experiments are briefed as follows:

1) FERET I (frontal) contains a gallery set: FA, with only a single image for each subjects (1,196 images in total); and 4 probe sets: FB, FC, DUP I, and DUP II (1,195, 194, 722, and 234 images, respectively), with strictly regulated facial expression, illumination, and time span variations [9]. Each image is re-aligned and cropped into 128 ×128 pixels referring to the annotated eye coordinates.

2) FERET II (non-frontal) is composed of 3 frontal reference gallery sets: BA, BJ and BK; and 6 non-frontal probe sets: BC, BD, BE, BF, BG, and BH, captured from viewpoint angles of ±40$^0$, ±25$^0$, ±15$^0$ [39]. Each of these repositories furnishes 200 images from the similar 200 subjects. As in FERET I, each image is geometrically pre-processed into 128 × 128 pixels, with respect to the manually annotated eyes and mouth coordinates.

3) AR encloses 2,600 frontal faces provided by 100 subjects in two contact sessions [40]. Each subject contributes 26 images of numerous expressions: neutral, smile, anger, scream; illumination modes: left or/and right lighting on; and occlusions: either sun-glasses or scarf. For each subject, only 2 out of 6 images with neutral expression are on the gallery list, and the remaining 24 images serve as probes. Each image is of 165 × 120 pixels in our experiments.

4) LFW is an unconstrained face verification dataset with 13,233 images from 5,749 subjects. Besides LFW-a [42], the normalized LFW-HPEN [36] dataset is also considered, where poses and expressions are eliminated with respect to the pre-learned 3D Morphable model. Some exemplars for the original LFW and the LFW-HPEN images are depicted in Fig. 7. The LFW-a and the re-aligned LFW-HPEN images1 are of each 150 × 80 and 88 × 64 pixels, respectively, in the 10-fold cross-validation setting. The six standard LFW evaluation settings are detailed in [43]. This paper, however, only considers the unsupervised, and the unrestricted, label-free outside data protocols.

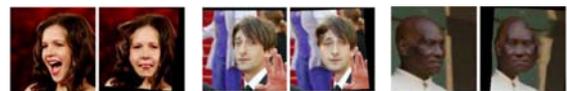

Fig. 7. Original and HPEN-normalized LFW images.

5) YouTube Faces (YTF) [57], i.e., the LFW celebrity subset accompanying with the LFW 10-fold evaluation protocol, consists of 3,425 online videos of 1,595 subjects sourced from YouTube for large-scale challenging, unconstrained face recognition in videos. On average, each subject owns 2.15 videos, with the shortest clip length of 48 frames, the longest of 6,070 frames, and a mean of 181.3 frames. The YTF video collection is of amateur photography, typically under problematic lighting, poses, occlusions, motion blur, etc., and also artifacts as a consequence of compression. The aligned frame images[1] are cropped into $120 \times 100$ each for feature extraction and encoding.

### B. Implementation Summary

Prior to the 2-FFC filter diversification stage, a set of 500,000 random mean-removed patches are sampled from the training (or gallery) images available. This is to learn the PCA and the ICA filter ensembles for each dataset, except for LFW-HPEN, as the undesirable artifacts imposed on the LFW-HPEN images (resulted from the pose and expression normalization) may lead the learned filters to be noisy. Therefore, the filter ensembles for LFW-HPEN are designated with the FERET-learned PCA and ICA filters. In our experiments, each PCA and ICA filter ensemble consists of 8 filters for all datasets.

In a nutshell, to formulate the 2-FFC$_{Gabor}$ descriptor, each image is convolved with the 36 unique offspring reposited in $\mathcal{O}_{Gabor}$. The filter responses are zero-thresholded and encoded into 8 feature images denoted by 8-bit integers. Each feature image is regionalized into $\mathcal{B}$ non-overlapping blocks, unless stated otherwise, and the block-wise histograms are cascaded into 8 histograms, one for each feature image. Subsequent to that, the histogram features for each real and imaginary part are aggregated to define the holistic histogram representation. The 2-FFC$_{Gabor}$ descriptor for an FERET I image of $8 \times 8$ blocks, as an example, is of 262,144 dimensions: $(8 \times 8 \text{ blocks}) \times 2^8$ bins $\times 8$ histograms $\times 2$, to be halved into 131,072 dimensions using the default **POOL$_{Avg}$** operator. This is followed by the square root and the L2 normalization operations, and the final WPCA compression to obtain the 2-FFC$_{Gabor}$ features for performance evaluation based on the nearest neighbor (NN) classifier with Cosine similarity scores.

The derivation procedures for the 2-FFC$_{Gabor+PCA}$ and the 2-FFC$_{Gabor+ICA}$ descriptors, in general, are equivalent to that of 2-FFC$_{Gabor}$, unless $\mathcal{O}_{Gabor-PCA}$ and $\mathcal{O}_{Gabor-ICA}$ are applied in the convolutional stage. The experimental results presented in the following sections are obtained with respect to the parameters configured in Table I.

### C. Performance Analysis

This section employs FERET I as a testbed to investigate the **Gabor$_{Cond}$** and **POOL$_{Avg}$** performance. The single-fold Gabor, PCA and ICA descriptors, i.e., 1-FFC$_{Gabor}$, 1-FFC$_{PCA}$, and 1-FFC$_{ICA}$, are also explored to unveil the $\mathcal{M}$-FFC benefit. The empirical results in this section are obtained based on the 1,000-dimensional WPCA-compressed features for PCANet [14], BSIF [17], BSIF$_{Face}$ [18], and DCTNet [20].

#### 1) Performance Analysis on Gabor$_{Cond}$

The first-line evaluation is to analyze the **Gabor$_{Cond}$** filter set and compare its performance to that of **Gabor$_{Std}$**. In this section, 1-FFC$_{Gabor}$ and 1-FFC$_{Gabor-40}$ correspond to the 1-FFC descriptors derived based on the 8 **Gabor$_{Cond}$** filters and the 40 **Gabor$_{Std}$** filters, respectively. While the 1-FFC$_{Gabor}$ descriptor is denoted by a single holistic histogram encoding the features conveyed by 8 filter responses, the 1-FFC$_{Gabor-40}$ descriptor is a concatenation of 5 histograms, one for each spatial frequency pre-tuned during the **Gabor$_{Std}$** construction. The 1-FFC$_{Gabor-40}$ descriptor, in fact, resembles the GGPP representation in [25]. What distinguish the 1-FFC$_{Gabor-40}$ features from GGPP is that the former compresses the demodulated Gabor phase features into 1,000 dimensions using WPCA. The rank-1 identification rates (%) in Table II validate our practice substituting **Gabor$_{Std}$** with **Gabor$_{Cond}$**. This is reasoned by the frequency spectrums depicted in Fig. 2.

TABLE I
PARAMETER CONFIGURATION FOR FERET I, FERET II, AR, LFW AND YTF, WHERE $^+$ DENOTES OVERLAPPING IMAGE REGIONALIZATION (INSTEAD OF NON-OVERLAPPING) TO COPE WITH POSE VARIATION.

| DATASET | # BLK, $\mathcal{B}$ | POOL$_{AVG}$ | FEA. DIM. BF. / AF. WPCA |
|---|---|---|---|
| FERET I | $8 \times 8$ | $P = S = 2$ | 131,072 / 1,000 |
| FERET II | $^+8 \times 8$ | $P = S = 2$ | 131,072 / 300 |
| AR | $8 \times 8$ | $P = S = 2$ | 131,072 / 180 |
| LFW-a (Unsupervised) | $^+10 \times 6$ | $P = S = 2$ | 122,880 / 2,000 |
| LFW-HPEN (Unsupervised) | $^+11 \times 8$ | $P = S = 4$ | 90,112 / 2,000 |
| LFW-HPEN (Unrestricted JB) | $^+11 \times 8$ | $P = S = 4$ | 90,112 / 500 |
| YTF (Restricted) | $^+8 \times 6$ | $P = S = 2$ | 98,304 / 2,000 |

TABLE II
PERFORMANCE SUMMARY FOR 1-FFC$_{GABOR}$ OF 8 FILTERS AND 1-FFC$_{GABOR-40}$ OF 40 FILTERS, IN TERMS OF RANK-1 IDENTIFICATION RATE (%).

| DESCR. | FB | FC | DUP I | DUP II | MEAN |
|---|---|---|---|---|---|
| 1-FFC$_{Gabor-40,7\times7}$ | 99.16 | 98.97 | 91.69 | 88.03 | 94.46 |
| 1-FFC$_{Gabor-40,9\times9}$ | 99.00 | 98.97 | **92.52** | **89.32** | **94.95** |
| 1-FFC$_{Gabor,7\times7}$ | 99.16 | **99.48** | 91.83 | 86.75 | 94.31 |
| 1-FFC$_{Gabor,9\times9}$ | **99.25** | 98.97 | 91.41 | **87.61** | 94.31 |

#### 2) Performance Analysis on POOL$_{Avg}$

The default **POOL$_{Avg}$** operator, with $P = S = 2$ as in (17), is investigated. It strides over every two successive histogram elements to averagely halve the full-dimensional 2-FFC$_{Gabor,Full}$ features, from 262,144 to 131,072 dimensions. The histogram distribution in Fig. 8 (a) depicts that the 2-FFC$_{Gabor,Full}$ features concentrate on some particular bins, leaving with a number of empty bins. This disparity would probably lead to biasness in the Cosine similarity score estimation. Fig. 8 (b), on the other hand, discloses that the histogram distribution is regulated and the empty bins are mostly sealed after the **POOL$_{Avg}$** operation. This evidences why the truncated histogram representation does not seem to undermine the overall performance (refer to Table III).

In addition to **POOL$_{Avg}$**, the max-pooling **POOL$_{Max}$** operator is probed. We perceive from Table III that **POOL$_{Avg}$** marginally outperforms **POOL$_{Max}$**. Employing **POOL$_{Max}$** to the histogram features deforms the original representation. Other informative components, except the maximum one, are discarded owing to dimension reduction.

---

[1] http://www.cslab.openu.ac.il/download/

TABLE III
PERFORMANCE SUMMARY FOR 2-FFC$_{\text{Gabor}}$ DESCRIPTORS WITH / WITHOUT **POOL$_{\text{Avg}}$**, AND **POOL$_{\text{Max}}$**, IN TERMS OF RANK-1 IDENTIFICATION RATE (%).

| DESCR. | POOL OP. | FB | FC | DUP I | DUP II | MEAN | FEA. DIM. |
|---|---|---|---|---|---|---|---|
| 2-FFC$_{\text{Gabor,Full,13×13}}$ | - | 99.33 | **100** | 95.57 | 93.59 | 97.12 | 262,144 |
| 2-FFC$_{\text{Gabor,13×13}}$ | POOL$_{\text{Avg}}$ | 99.41 | **100** | **95.98** | **94.02** | **97.35** | 131,072 |
| 2-FFC$_{\text{Gabor,13×13}}$ | POOL$_{\text{Max}}$ | 99.33 | **100** | 95.71 | 93.59 | 97.16 | 131,072 |

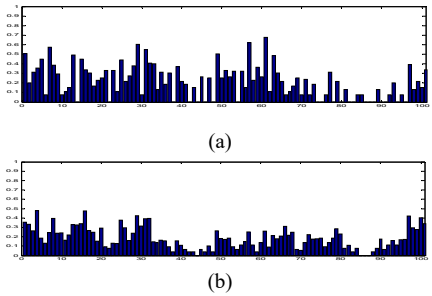

Fig. 8. Histogram distribution for 2-FFC$_{\text{Gabor}}$ (a) without **POOL$_{\text{Avg}}$**; (b) with **POOL$_{\text{Avg}}$**.

### 3) Performance Analysis on 1-FFC and 2-FFC

Besides performance investigation on the 1-FFC and the 2-FFC descriptors, the ICA filter learning practices in [17] and [18] are studied. Our experiments re-use the natural images-learned ICA filters shared by the authors [17]; and the block-wise ICA filters in BSIF$_{\text{Face}}$ [18] is trained for each face region from the FERET-FA images. The key findings perceived from Table IV are:

i) The performance comparison of the 1-FFC$_{\text{ICA}}$ and the BSIF descriptors disclose that the face images-trained ICA filters outperform the natural images-learned counterparts. The 1-FFC$_{\text{ICA}}$ performance, however, is far from satisfactory, as compared to other face descriptors (to be discussed in next section).

ii) In comparison with that of holistic ICA filters, the BSIF$_{\text{Face}}$ block-wise ICA filters are of far-fetched. This is probably due to the over-fitting problem, although the proposed soft assignment (SA) smoothing normalization is also applied.

iii) For the 2-FFC$_{\text{Gabor-PCA}}$ and the 2-FFC$_{\text{Gabor-ICA}}$ instances, the vanishing of the orthonormal and independent traits of the PCA and ICA filters (due to 2-FFC) is compensated with at least 3% of improvement. This proves that the number of convolutional filters is also a critical performance factor.

Note that, the 1-FFC$_{\text{PCA}}$ construction is analogous to that of 1-layer PCANet, codenamed 1-PCANet in Table IV. The TT pre-processing step [19] is simply withdrawn from the BSIF and the BSIF$_{\text{Face}}$ pipelines, for a fair comparison.

### D. Performance Evaluation

This section presents the performance for the 2-FFC$_{\text{Gabor}}$, the 2-FFC$_{\text{Gabor-PCA}}$, and the 2-FFC$_{\text{Gabor-ICA}}$ descriptors with respect to the aforementioned benchmarking datasets.

TABLE IV
PERFORMANCE SUMMARY FOR 1-FFC, 2-FFC, BSIF, BSIF$_{\text{FACE}}$ AND 1-PCANET DESCRIPTORS, IN TERMS OF RANK-1 IDENTIFICATION RATE (%). FOR A FAIR COMPARISON, ALL FACE DESCRIPTORS ARE WPCA-COMPRESSED INTO 1,000 DIMENSIONS.

| DESCR. | FB | FC | DUP I | DUP II | MEAN |
|---|---|---|---|---|---|
| BSIF$_{7×7}$ [17] (ICPR 2012) | 99.33 | 98.45 | 90.17 | 85.47 | 93.36 |
| BSIF$_{\text{Face},7×7}$ [18] (SCIA, 2015) | 98.16 | 99.48 | 89.75 | 86.75 | 93.54 |
| BSIF$_{\text{Face},7×7}$ + SA [18] (SCIA, 2015) | 98.33 | 99.48 | 88.23 | 85.04 | 92.77 |
| 1-PCANet$_{7×7}$ [14] (TIP, 2015) | 99.08 | 98.97 | 91.14 | 87.18 | 94.09 |
| 1-FFC$_{\text{Gabor},7×7}$ | 99.16 | 99.48 | 91.83 | 86.75 | 94.31 |
| 1-FFC$_{\text{PCA},7×7}$ | 99.08 | 98.97 | 91.14 | 87.18 | 94.09 |
| 1-FFC$_{\text{ICA},7×7}$ | 99.25 | 98.45 | 92.52 | 88.03 | 94.56 |
| 2-FFC$_{\text{Gabor},13×13}$ | 99.41 | **100** | 95.98 | 94.02 | 97.35 |
| 2-FFC$_{\text{Gabor-PCA},13×13}$ | **99.50** | **100** | 96.26 | **94.87** | 97.66 |
| 2-FFC$_{\text{Gabor-ICA},13×13}$ | **99.50** | **100** | 96.12 | **94.87** | 97.62 |

### 1) Face Identification: FERET I

The three 2-FFC descriptors are analyzed and compared to the recent face descriptors, in terms of rank-1 identification rate (%), in Table V. On average, the 2-FGFC descriptors prevail over the conventional 1-fold Gabor descriptors: HGPP [25], and G-LQP [28]. The merit of the 2-FFC descriptors is that they are directly derived from the 2-FFC offspring sets for each real and imaginary part, without any non-linearity like LBP or LXP. In the meantime, G-LQP histograms the LBP-manipulated Gabor patterns based on a pre-learned codebook.

PCANet [14] are re-implemented using the FERET-learned PCA filters. Other notable learning-based face descriptors, e.g., the LDA-based LBP variants: DFD [15] and CBFD [16] (the filters are learned from the LBP-alike neighborhood properties based on the supervised LDA criterions); the two BSIF derivatives: BSIF [17] (use of the natural image-learned ICA filters) and BSIF$_{\text{Face}}$ [18] (use of the block-based ICA filters); and the learning-free DCTNet [20] are also selected for performance comparison. Besides of being parsimonious, the data-independent 2-FFC$_{\text{Gabor}}$ and the DCTNet descriptors are of more discriminative than all other learning-based descriptors, especially on the most challenging DUP I and DUP II probe sets. However, the DCTNet features are regularized via the proposed tied-rank (TR) normalization. The inclusion of the PCA and ICA filters in feature extraction further improves the 2-FFC performance, marking an average rank-1 identification rate of 97.66% for the 2-FFC$_{Gabor-PCA}$ descriptor, and 97.62% for the 2-FFC$_{Gabor-ICA}$ descriptors. Another finding is that the two BSIF descriptors rely on the TT normalization [19]. Their performance drops drastically, from 96.19% to 93.36% (for BSIF), and from 95.65% to 93.54% (for BSIF$_{Face}$), without TT.

TABLE V
PERFORMANCE SUMMARY FOR 2-FFC AND STATE OF THE ART DESCRIPTORS, IN TERMS OF RANK-1 IDENTIFICATION RATE (%), ON **FERET I**, WHERE * DENOTES RESULTS REPORTED IN ORIGINAL PAPERS. FOR A FAIR COMPARISON, 2-FFC, PCANET, DCTNET, BSIF AND BSIF$_{FACE}$ DESCRIPTORS ARE WPCA-COMPRESSED INTO 1,000 DIMENSIONS.

| DESCR. | FB | FC | DUP I | DUP II | MEAN |
|---|---|---|---|---|---|
| * HGPP [25] (TIP, 2007) | 97.50 | 99.50 | 79.50 | 77.80 | 88.58 |
| * G-LQP [28] (BCMV, 2012) | **99.99** | **100** | 93.20 | 91.00 | 96.05 |
| BSIF$_{7\times7}$ [17] + TT [19] (ICPR 2012) | 99.08 | **100** | 93.35 | 92.31 | 96.19 |
| BSIF$_{7\times7}$ [17] (ICPR 2012) | 99.33 | 98.45 | 90.17 | 85.47 | 93.36 |
| PCANet$_{FERET,5\times5}$ [14] (TIP, 2015) | 99.16 | **100** | 94.04 | 92.31 | 96.38 |
| * DFD [15] (TPAMI, 2014) | 99.40 | **100** | 91.80 | 92.30 | 95.88 |
| * CBFD [16] (TPAMI, 2015) | 99.80 | **100** | 93.50 | 93.20 | 96.63 |
| BSIF$_{Face,7\times7}$ [18] + TT [19] (SCIA, 2015) | 98.41 | **99.48** | 92.38 | 92.31 | 95.65 |
| BSIF$_{Face,7\times7}$ [18] (SCIA, 2015) | 98.16 | 99.48 | 89.75 | 86.75 | 93.54 |
| DCTNet$_{5\times5}$ [20] (APSIPA, 2015) | 99.08 | **100** | 93.35 | 91.45 | 95.97 |
| DCTNet$_{5\times5}$ + TR [20] (APSIPA, 2015) | 99.67 | **100** | 95.57 | 94.02 | 97.32 |
| 2-FFC$_{Gabor,13\times13}$ | 99.41 | **100** | 95.98 | 94.02 | 97.35 |
| 2-FFC$_{Gabor-PCA,13\times13}$ | 99.50 | **100** | **96.26** | **94.87** | **97.66** |
| 2-FFC$_{Gabor-ICA,13\times13}$ | 99.50 | **100** | 96.12 | **94.87** | 97.62 |

*2) Face Identification: FERET II*

The 2-FFC descriptors are investigated against pose variation on the FERET II dataset, where the WPCA projection matrix is trained from the frontal gallery images. Table VI discloses that most of the descriptors are sensitive to the awful pose angles of $\pm 40^0$. Rather than non-overlapping blocks, the 2FFC descriptors are derived from the overlapping blocks such that each block occupies a broader histogramming region. The overlapping ratio is set to 0.5, where a non-overlapping block of $8 \times 8$ pixels is extended to $16 \times 16$, $16 \times 16$ to $32 \times 32$, and the like, with a step size of 8 and 16, respectively. This trick improves the 2-FFC performance by at least at 2%. We only compare the 2-FFC descriptors to the re-implemented PCANet and DCTNet due to the reason that there are limited papers reporting on this dataset. As a whole, the 2-FFC performance surpasses PCANet and DCTNet, as the 2-FFC descriptors are benefited from the 2-fold offspring, in conjunction with the overlapping trick.

TABLE VI
PERFORMANCE COMPARISON FOR 2-FFC, PCANET AND DCTNET DESCRIPTORS, IN TERMS OF RANK-1 IDENTIFICATION RATE (%), ON **FERET II**, WHERE $^+$ DENOTES 2-FFC DESCRIPTORS DEFINED BASED ON OVERLAPPING BLOCKS. FOR A FAIR COMPARISON, ALL FACE DESCRIPTORS ARE WPCA-COMPRESSED INTO 300 DIMENSIONS.

| DESCR. | B$_C$ +40$^0$ | B$_D$ +25$^0$ | B$_E$ +15$^0$ | B$_F$ -15$^0$ | B$_G$ -25$^0$ | B$_H$ -40$^0$ | MEAN |
|---|---|---|---|---|---|---|---|
| PCANet$_{5\times5}$ [14] (TIP, 2015) | 82.50 | 98.50 | 99.50 | **100** | 94.50 | 64.00 | 89.83 |
| DCTNet$_{5\times5}$ [20] (APSIPA, 2015) | 78.00 | **100** | **100** | **100** | 96.50 | 70.00 | 90.75 |
| DCTNet$_{5\times5}$ +TR [20] (APSIPA, 2015) | 86.50 | **100** | **100** | **100** | 99.50 | 88.00 | 95.67 |
| 2-FFC$_{Gabor, 17\times17}$ | 88.00 | **100** | **100** | **100** | 98.50 | 75.00 | 93.58 |
| $^+$2-FFC$_{Gabor, 17\times17}$ | 91.00 | **100** | **100** | **100** | 98.50 | 81.50 | 95.17 |
| $^+$2-FFC$_{Gabor-PCA, 17\times17}$ | 93.00 | **100** | **100** | **100** | 99.00 | 91.00 | 97.17 |
| $^+$2-FFC$_{Gabor-ICA, 17\times17}$ | **97.00** | **100** | **100** | **100** | 99.00 | **93.00** | **98.17** |

*3) Face Identification: AR*

The rank-1 identification rates (%) for the PCA-whitened 2-FFC, PCANet and DCTNet descriptors of 180 dimensions are reported in Table VII. The 2-FFC descriptors, on the whole, exhibit remarkable robustness over the PCANet and DCTNet descriptors, particularly for facial expressions, and sun-glasses and scarf disguises. It is noteworthy to accentuate that we only include 2 frontal faces with neutral expression per subject as the gallery references.

TABLE VII
PERFORMANCE COMPARISON FOR 2-FFC, PCANET AND DCTNET DESCRIPTORS, IN TERMS OF RANK-1 IDENTIFICATION RATE (%), ON **AR**. FOR A FAIR COMPARISON, ALL FACE DESCRIPTORS ARE WPCA-COMPRESSED INTO 180 DIMENSIONS.

| DESCR. | EXPR. | ILLM. | OCCL. | MEAN |
|---|---|---|---|---|
| PCANet$_{5\times5}$ [14] (TIP, 2015) | 94.28 | **100** | 97.90 | 97.39 |
| DCTNet$_{5\times5}$ [20] (APSIPA, 2015) | 94.11 | **100** | 97.64 | 97.25 |
| DCTNet$_{5\times5}$ + TR [20] (APSIPA, 2015) | 97.81 | **100** | 99.24 | 99.02 |
| 2-FFC$_{Gabor,17\times17}$ | 98.65 | **100** | **99.92** | 99.52 |
| 2-FFC$_{Gabor-PCA,17\times17}$ | 98.82 | **100** | 99.75 | 99.52 |
| 2-FFC$_{Gabor-ICA,17\times17}$ | **99.33** | **100** | **99.92** | **99.75** |

*4) Face Verification: LFW*

The first experiment examines the 2-FFC descriptors of 2,000 dimensions (after WPCA) on the LFW unsupervised evaluation protocol. Apart from AUC, we also provide ACC± SD, i.e., the best accuracy measured based on true positive rates (TPR) and false positive rates (FPR). The flip-free augmentation in [52] is practiced, such that the final score is obtained by averaging the Cosine similarity scores for the 4 combination pairs stemmed from the original and the horizontally flipped images. We discern from Table VIII that:

i) The 2-FFC descriptors perform equally well in face verification with an average AUC of above 94% for both LFW-a and LFW-HPEN.
ii) The 2-FFC performance is positioned in the second place, after MRF-Fusion-CSKDA [46]. Unlike our single-type 2-FFC descriptors, the MRF-Fusion-CSKDA descriptor is of multi-type cascading multi-scale LBP, multi-scale local phase quantization (LPQ), and multi-scale BSIF features. Instead of the naivest NN classifier, the MRF-Fusion-CSKDA pipeline recruits a nonlinear binary class-specific kernel discriminant analysis classifier.
iii) In place of one-to-one matching in our case, Spartans [47] synthesizes a number of new faces under a wide range of 3D rotations to yield discriminative matching scores. The Spartans performance, however, is capped at 94.28%, marginally below the 2-FFC descriptors, either on LFW-a, or LFW-HPEN.

The ROC curves for the 2-FFC descriptors and the most recent descriptors are portrayed in Fig. 9. Note that, the TPR and the FPR for MRF-Fusion-CSKDA are not provided.

The second experiment evaluates the 2-FFC performance on the LFW unrestricted, with label-free outside data protocol. As the class identity is made known to this setting, the supervised Joint Bayesian (JB) metric learning approach [48] is exercised, without any external training images (but only flipped images). This restriction forces us to reduce the 2-FFC features with only 500 dimensions after WPCA [49]. Instead of the NN classifier, the 2-FFC performance is probed with linear SVM. According to Table IX, our observations are:

i) The 2-FFC performance is comparable to that of 3-layer hybrid ConvNet-RBM network [34]. However, as appose to the ConvNet-RBM architecture, the 2-FFC pipeline is of far more simplistic. For example, the ConvNet-RBM network represents a single input image with 12 color and grayscale face regions; each pair of a region is expanded to 8 modes to extract the local relational visual features.
ii) The MLBPH + MLPQH + MBSIFH [50] and CBFD [16] descriptors are multi-scale and multi-type: the former is a combination of multi-scale LBP, LPQ, and BSIF (either in 4 or 5 scales); while the latter stacks the CBFD descriptors with another 5 descriptors, including LBP, etc. The 2-FFC descriptors are disclosed to be parallel to each individual MLBPH, MLPQH and MBSIFH. Furthermore, the 2-FFC descriptors are of more discriminative as compared to the single-input, single-type CBFD descriptors.
iii) The multi-scale HD-LBP [49], HPEN + HD-LBP [36], and HPEN + HD-Gabor [36] descriptors are derived from the dense fiducial landmarks detected by using Cao et al. [53]. The MDML-DCPs [31] descriptor is of multi-input, where its final performance is measured on a fusion score of 9 feature representations of a single image, including 2 holistic and 7 local landmark features.
iv) Another interesting remark is that most of the non-CNN descriptors, including the 2-FFC descriptors, outshines the LFW-trained deep CNN, i.e., MatConvNet + JB [51], even its performance is summarized from the best-performing 16 networks (out of 30). This attests that a deep-network would be underperformed, if the training examples are of insufficient.

TABLE VIII
PERFORMANCE COMPARISON FOR 2-FFC DESCRIPTORS AND STATE OF THE ARTS, IN TERMS OF **AUC** (%), AND **ACC ± SD** (%), UNDER **LFW UNSUPERVISED EVALUATION PROTOCOL**. NOTE THAT, * DENOTES THE EMPIRICAL RESULTS REPORTED IN THE RESPECTIVE PAPERS.

| DESCR. | AUC | ACC (%) ± SD | REMARK |
|---|---|---|---|
| * PAF [44] (CVPR, 2013) | 94.05 | - | |
| * MRF-MLBP [45] (BTAS, 2013) | 89.84 | - | |
| * MRF-Fusion-CSKDA [46] (TIFS, 2014) | **98.94** | - | |
| * Spartans [47] (TIP, 2015) | 94.28 | - | LFW-a |
| * G-LQP [28] (BCMV, 2012) | - | 82.10 ± 0.26 | |
| * PCANet$_{7\times7}$ [14] (TIP, 2015) | - | 86.28 ± 1.14 | |
| * DFD [15] (TPAMI, 2014) | - | 84.02 ± 0.40 | |
| * CBFD [16] (TPAMI, 2015) | 90.91 | - | |
| * BSIF$_{Face}$ [18] (SCIA, 2015) | 93.18 | - | |
| 2-FFC$_{Gabor,17\times17}$ | 93.30 | 87.37 ± 1.46 | |
| 2-FFC$_{Gabor-PCA,17\times17}$ | 94.59 | 88.63 ± 1.40 | LFW-a |
| 2-FFC$_{Gabor-ICA,17\times17}$ | 94.36 | 88.18 ± 1.18 | |
| 2-FFC$_{Gabor,17\times17}$ | 95.88 | 90.67 ± 1.35 | |
| 2-FFC$_{Gabor-PCA,17\times17}$ | 95.54 | 90.00 ± 1.09 | LFW-HPEN |
| 2-FFC$_{Gabor-ICA,17\times17}$ | 95.89 | 90.53 ± 1.10 | |

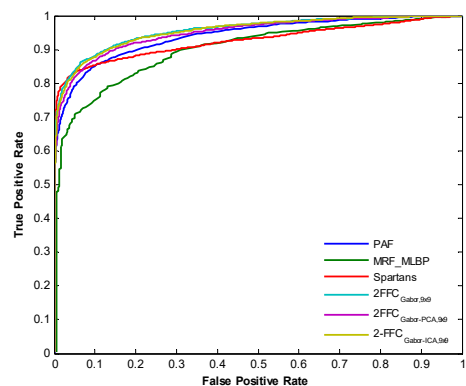

Fig. 9. Comparison summary for 2-FFC descriptors and most recent state of the arts, in terms of ROC averaged over 10 folds for **LFW** unsupervised evaluation protocol.

To conclude, the existing face descriptors are leveraged by taking multiple representations of a single image for feature extraction, either in multiple scales, or/and in multiple regions. Besides, the multi-type descriptors, e.g., [16] and [50], do not reflect the actual performance, and therefore unreasonable to compare with that of single-type and single-input such as the 2-FFC descriptors. For performance gain, it is possible to cascade the 2-FFC descriptors with the features extracted from different image representations, and other descriptors of varying types.

TABLE IX
PERFORMANCE COMPARISON FOR 2-FFC DESCRIPTORS AND STATE OF THE ARTS, IN TERMS OF **ACC ± SD** (%), ON **LFW**, UNDER **LFW UNRESTRICTED**, LABEL-FREE OUTSIDE DATA EVALUATION PROTOCOL. NOTE THAT, * INDICATES THE EXPERIMENTAL RESULTS REPORTED IN THE RESPECTIVE PAPERS.

| DESCR. | ACC (%) ± SD | REMARK |
|---|---|---|
| * HD-LBP [49] + JB (CVPR, 2013) | 93.18 ± 1.07 | Multi-Scale Single-Type |
| * ConvNet-RBM [34] (ICCV, 2013) | 91.75 ± 0.48 | Multi-Input Single-Type |
| * MLBPH [50] (ICIP, 2014) | 88.60 ± 0.87 | Multi-Scale Single-Type |
| * MLPQH [50] (ICIP, 2014) | 91.27 ± 0.71 | Multi-Scale Single-Type |
| * MBSIFH [50] (ICIP, 2014) | 91.40 ± 0.94 | Multi-Scale Single-Type |
| * MLBPH + MLPQH + MBSIFH [50] (ICIP, 2014) | 93.03 ± 0.82 | Multi-Scale Multi-Type |
| * CBFD [16] + JB (TPAMI, 2015) | 87.87 ± 1.86 | Single-Input (a) Single-Type |
| | 88.90 ± 1.81 | Single-Input (b) Single-Type |
| | 88.35 ± 1.61 | Single-Input (c) Single-Type |
| | 90.90 ± 1.40 | Multi-Input (a, b, c) Single-Type |
| | 93.80 ± 1.31 | Multi-Input (a, b, c) Multi-Type |
| * HPEN + HD-LBP [36] + JB (CVPR, 2015) | 94.87 ± 0.38 | Multi-Scale Single-Type |
| * HPEN + HD-Gabor [36] +JB (CVPR, 2015) | 95.25 ± 0.36 | |
| * MDML-DCPs + JB [31] + Linear SVW (TPAMI, 2015) | **95.40 ± 0.33** | Multi-Input Single-Type |
| * MatConvNet [51] + JB ([cs.CV], 2015) | 87.63 ± 0.64 | Multi-Scale Multi-Input |
| 2-FFC$_{\text{Gabor},17\times17}$ + JB + Linear SVM | 91.43 ± 1.09 | Single-Input Single-Type |
| 2-FFC$_{\text{Gabor-PCA},17\times17}$ + JB + Linear SVM | 91.81 ± 1.79 | |
| 2-FFC$_{\text{Gabor-ICA},17\times17}$ + JB + Linear SVM | 91.57 ± 1.63 | |

TABLE X
PERFORMANCE COMPARISON FOR 2-FFC DESCRIPTORS AND STATE OF THE ARTS, IN TERMS OF **ACC ± SD** (%), ON **YTF**, UNDER **RESTRICTED EVALUATION PROTOCOL**. NOTE THAT, * INDICATES THE EXPERIMENTAL RESULTS REPORTED IN THE RESPECTIVE PAPERS.

| DESCR. | ACC (%) ± SD |
|---|---|
| * MBGS - LBP [57] (CVPR, 2011) | 76.40 ± 1.80 |
| * PHL - LBP + SILDA [59] (TIP, 2013) | 80.20 ± 1.30 |
| * DeepFace-Single [3] (CVPR, 2014) | **91.40 ± 1.10** |
| * DDML - LBP + CSLBP + FPLBP [60] (CVPR, 2014) | 82.34 ± 1.47 |
| * EigenPEP [61] (ACCV, 2014) | 84.80 ± 1.40 |
| * LM3L - LBP + CSLBP + FPLBP [62] (ACCV, 2014) | 81.30 ± 1.20 |
| 2-FFC$_{\text{Gabor},17\times17}$ + SILDA | 85.12 ± 1.26 |
| 2-FFC$_{\text{Gabor-PCA},17\times17}$ + SILDA | **86.60 ± 1.17** |
| 2-FFC$_{\text{Gabor-ICA},17\times17}$ + SILDA | 86.34 ± 1.12 |

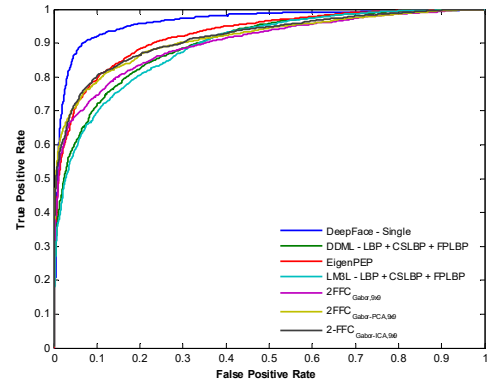

Fig. 10. Comparison summary for 2-FFC descriptors and most recent state of the art, in terms of ROC averaged over 10 folds on **YTF**.

*5) Face Verification: YTF*

Following the restricted protocol, we validate the video-level face verification performance for the 2-FFC descriptors using the YTF dataset. We select a random frame from each video, i.e., a summation of 3,425 frames, for the PCA and ICA filter ensemble construction. The flip-free strategy used in the LFW evaluation is again borrowed to augment the frames available for WPCA training. Subsequent to that, the weakly-supervised side-information LDA (SILDA) in [58] is implemented based on the same and not-same pair information. Given a test pair, a number of $r$ frames are randomly selected from each video for feature extraction, followed by WPCA and SILDA dimension reduction. For each fold, ACC is measured based on the average over $4 \times r$ matching scores. We set $r$ to 20 in our experiments, and the 2-FFC descriptors are compressed to 2,000 dimensions using WPCA and SLIDA.

The overall 2-FFC performance is compared to the baseline descriptors in Table X, and the ROC curves are displayed in Fig. 10. We note that:

i) Although the 2-FFC descriptors are of naïve to video face recognition (as we only perform frame-to-frame matching and average the matching scores across the frame pairs), they are ranked second, after DeepFace - Single [3].

ii) The single-type 2-FFC descriptors remarkably outperform other two multi-type descriptors, namely, DDML - LBP + CSLBP + FPLBP [60], and LM3L - LBP + CSLBP + FPLBP [62], by at least 2% in AUC and 4% in ACC. Also note that the SILDA used by the 2-FFC descriptors is an ordinary linear LDA, while the DDML metric learning is of non-linear type with deep network structure.

iii) We also observe that the 2-FFC descriptors, particularly 2-FFC$_{\text{Gabor-PCA}}$ and 2-FFC$_{\text{Gabor-ICA}}$, are on a par with the EigenPEP video descriptor [61]. The 2-FFC descriptors, however, might not be a scalable solution as the complexity relies on the number of frames the video possesses.

*6) Discussion Summary*

As a whole, we summarize the 2-FFC performance as follows:

i) The empirical results disclose that, to derive a highly robust face descriptor, one can recruit more convolution filters for feature extraction. To confront with the filter scarcity issue, this work instantiates an adequate number of offspring by means of $\mathcal{M}$-FFC filter diversification.

ii) In a 2-FFC Gabor-Gabor offspring set, the Gabor-PCA and Gabor-ICA offspring are of more discriminative in feature extraction. This reveals that the reason why the 2-FFC$_{Gabor-PCA}$ and the 2-FFC$_{Gabor-ICA}$ descriptors consistently perform better than that of 2-FFC$_{Gabor}$.

iii) The 2-FFC descriptors have shown remarkable superiority over PCANet, BSIF, DCTNet, and other filter bank-based face descriptors, not only in the strictly regulated condition but also the challenging unconstrained settings. Although the 2-FFC architecture is merely a single-flat network, the 2-FFC performance is shown on a par with that of the 3-layer hybrid ConvNet-RBM [34]. This further certifies the discriminability of the $\mathcal{M}$-FFC descriptors.

iv) The impressive performance suggests to adopt the $\mathcal{M}$-FFC descriptors as a baseline for analyzing a more sophisticated models, e.g., CNNs or video-level face recognition models.

## V. CONCLUSION

This paper outlines a new means of filter diversification, where Gabor, PCA and ICA filters are cross convolved by $\mathcal{M}$-folds to instantiate offspring filter sets for feature extraction. We reveal through extensive experiments that it is viable to condense the 40 multi-scale, multi-orientation Gabor filters into only 8 filters for face recognition. The average histogram pooling operator, on the other hand, is disclosed to be indispensable in two roles: dimension reduction, and soft regularization. We reveal that the 2-FFC descriptors are superior to other face descriptors on three face identification datasets, i.e., FERET I (frontal), FERET II (non-frontal), and AR. We also show that the 2-FFC descriptors achieve the face verification standard scrutinized on LFW and video-level YTF. For future work, the single-type, single-input 2-FFC descriptors will be explored by considering multi-scale, or multi-input images. The $\mathcal{M}$-FFC downside will be remedied to bound the offspring set to a reasonable size, for $\mathcal{M} > 2$.

## VI. ACKNOWLEDGEMENT

This work was supported by the National Research Foundation of Korea (NRF) grant funded by the Korea government (MSIP) (NO. 2016R1A2B4011656).

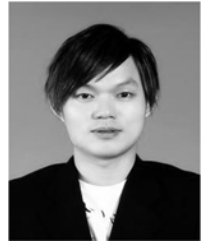

**Cheng Yaw Low** received the B.S. and M.S. degrees in information technology from Multimedia University (MMU), Malaysia, in 2004 and 2009, respectively. He is currently on study leave from MMU to pursue the Ph.D. degree at the School of Electrical and Electronic Engineering, Yonsei University, South Korea. His research interest includes machine learning and pattern recognition.

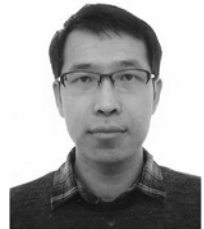

**Andrew Beng Jin Teoh (SM'12)** received the Ph.D. degree in electrical, electronic and system engineering from National University of Malaysia, in 2003. He is an Associate Professor with the School of Electrical and Electronic Engineering, Yonsei University, South Korea. He has published over 300 international refereed journals, conference articles, and several book chapters in the areas of biometrics and machine learning. He served and is serving as a guest editor of IEEE Signal Processing Magazine, associate editor of IEEE Biometrics Compendium and editor-in-chief of IEEE Biometrics Council Newsletter.

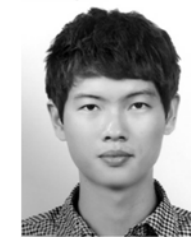

**Cong Jie Ng** received the B. S. degree in computer science from Multimedia University, Malaysia, in 2011. He is currently a Master candidate in School of Electrical and Electronic Engineering, Yonsei University, South Korea. His research interests are in computer vision and machine learning.